\documentclass[letterpaper, 10 pt, conference]{ieeeconf}
\IEEEoverridecommandlockouts                             
\usepackage[english]{babel}
\usepackage{hyperref}

\usepackage{cite}
\usepackage{amsmath}
\usepackage[pdftex]{graphicx}
\usepackage{epstopdf}

\title{\LARGE \bf LIDAR-Camera Fusion for Road Detection \\ Using Fully Convolutional Neural Networks}

\author{Luca Caltagirone$^{*}$, Mauro Bellone,
	Lennart Svensson, Mattias Wahde\\
	{\tt\small\{luca.caltagirone, mauro.bellone, lennart.svensson, mattias.wahde\}@chalmers.se}
	\thanks{$^{*}$Corresponding author. Luca Caltagirone, Mauro Bellone, and Mattias Wahde are with the Adaptive Systems Research Group, Department of Mechanics and Maritime Sciences, Chalmers University of Technology, Gothenburg, Sweden.
Lennart Svensson is with the Department of Electrical Engineering, also at Chalmers University of Technology.}}

\begin{document}

\maketitle
\thispagestyle{empty}
\pagestyle{empty}

\begin{abstract}
In this work, a deep learning approach has been developed to carry out road detection by fusing LIDAR point clouds and camera images. An unstructured and sparse point cloud is first projected onto the camera image plane and then upsampled to obtain a set of dense 2D images encoding spatial information. Several fully convolutional neural networks (FCNs) are then trained to carry out road detection, either by using data from a single sensor, or by using three fusion strategies: early, late, and the newly proposed \textit{cross fusion}. Whereas in the former two fusion approaches, the integration of multimodal information is carried out at a predefined depth level, the cross fusion FCN is designed to directly learn from data where to integrate information; this is accomplished by using trainable cross connections between the LIDAR and the camera processing branches.

To further highlight the benefits of using a multimodal system for road detection, a data set consisting of visually challenging scenes was extracted from driving sequences of the KITTI raw data set. It was then demonstrated that, as expected, a purely camera-based FCN severely underperforms on this data set. A multimodal system, on the other hand, is still able to provide high accuracy.
Finally, the proposed cross fusion FCN was evaluated on the KITTI road benchmark where it achieved excellent performance, with a MaxF score of 96.03\%, ranking it among the top-performing approaches. 
\end{abstract}

\section{Introduction}
% % % % context and background
Road detection is an important task that needs to be solved accurately and robustly in order to achieve higher automation levels. Knowing what regions of the road surface are available for driving is in fact a crucial prerequisite for carrying out safe trajectory planning and decision-making.
Although some automated driving vehicles are already available on the market, the recent crash of a Tesla car controlled by its autopilot system highlighted that further research and testing are very much necessary. In that case, it was pointed out that a possible reason for the crash was that the autopilot system misinterpreted the trailer of a truck as free road due to unfavourable lighting conditions \cite{EvanIEEEspec, Endsley17}.  

Current approaches for road detection use either cameras or LIDAR sensors. Cameras can work at high frame-rate, and provide dense information over a long range under good illumination and fair weather. However, being passive sensors, they are strongly affected by the level of illumination. 
A passive sensor is able to receive a specific amount of energy from the environment, light waves in the case of cameras, and transform it into a quantitative measure (image). 
Clearly, the process depends on the amplitude and frequency of the light waves, influencing the overall result, while a reliable system should be invariant with respect to changes in illumination \cite{Alvarez2007}.
LIDARs sense the environment by using their own emitted pulses of laser light and therefore they are only marginally affected by the external lighting conditions. Furthermore, they provide accurate distance measurements. However, they have a limited range, typically  between 10 and 100 meters, and provide sparse data. 

Based on this description of benefits and drawbacks of these two sensor types, it is easy to see that using both might provide an improved overall reliability. Inspired by this consideration, the work presented here investigates how LIDAR point clouds and camera images can be integrated for carrying out road segmentation.
The choice to use a fully convolutional neural network (FCN) for LIDAR-camera fusion is motivated by the impressive success obtained by deep learning algorithms in recent years in the fields of computer vision and pattern recognition \cite{LecunEtAl2015}.

In summary, this work makes the following two main contributions: 
(i) A novel LIDAR-camera fusion FCN that outperforms established approaches found in the literature and achieves state-of-the-art performance on the KITTI road benchmark;
(ii) a data set of visually challenging scenes extracted from KITTI driving sequences that can be used to further highlight the benefits of combining LIDAR data and camera images for carrying out road segmentation. 

The remainder of the paper is structured as follows:
Sect.~\ref{sec:related_work} gives a brief overview of related approaches that deal with the problems of road detection or sensor fusion. The FCN base architecture and the fusion strategies are presented in Sect.~\ref{sec:deepstuff}. Section \ref{sec:datapreprocessing} describes the procedure to transform a sparse 3D point cloud into a set of dense 2D images. The experimental results and discussion are reported in Sect.~\ref{sec:experimental_results} which is followed, in Sect.~\ref{sec:conclusion}, by a summary and the conclusions.

\begin{figure*}[h]
	\centering
	\includegraphics[width=\textwidth]{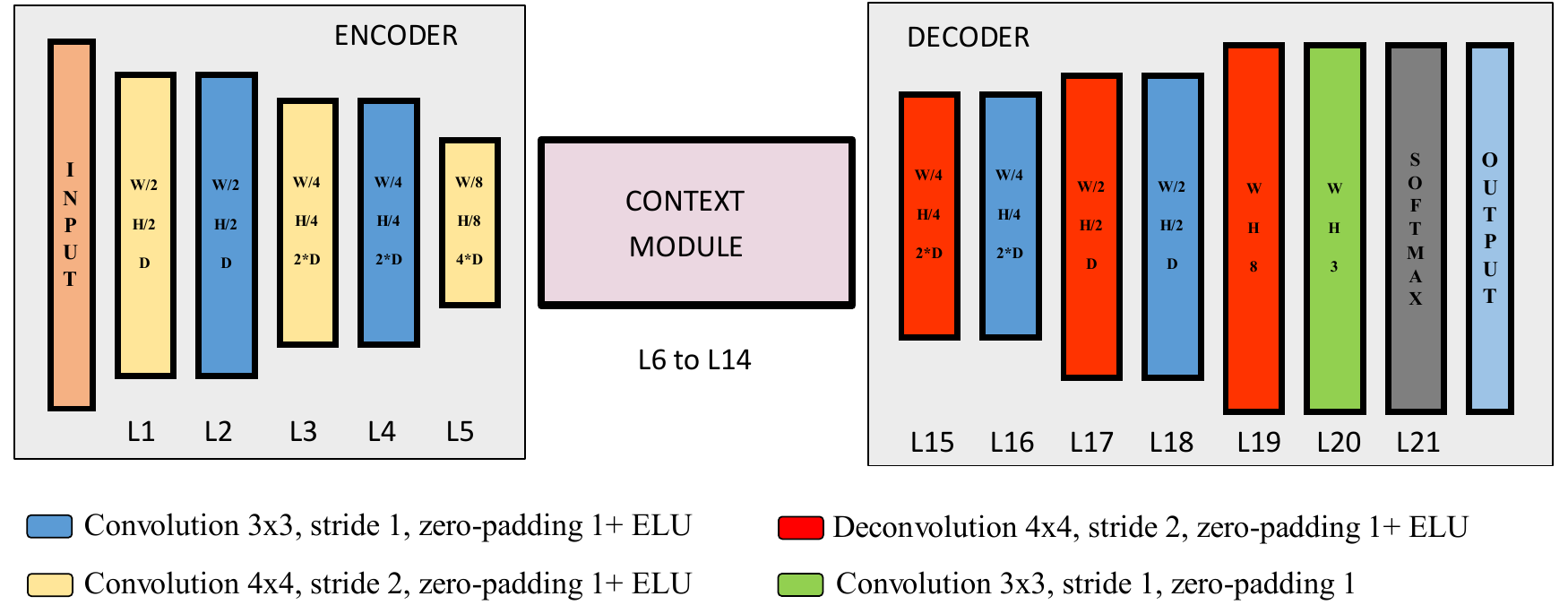}
	\caption{A schematic illustration of the proposed base FCN architecture which consists of 21 layers. W represents the width, H denotes the height, and D is the number of feature maps in the first layer which was set to 32. The FCN uses the exponential linear unit (ELU) activation function after each convolutional layer. See Table~\ref{tab:nets_properties} for details about the context module architecture.}
	\label{fig:network}
\end{figure*}

\section{Related work}
\label{sec:related_work}
	The study of road detection can be tracked back a few decades; 
	in \cite{Broggi95} and \cite{Broggi95_2}, Broggi \textit{et al.}~already presented 
	an algorithm for the binarization, classification, and interpretation of visual images for road detection.
	However, recent advances in sensor development and hardware computation capacity has made possible the use of 
	high-accuracy methods. 
	Nowadays, the large majority of state-of-the-art algorithms for road detection use, to different extent, machine learning techniques. 
	Teichmann \textit{et al.} \cite{TeichmannEtAl2016}, for example, trained a convolutional neural network (CNN) to simultaneously solve the tasks of road segmentation and vehicle detection in monocular camera images. Chen \textit{et al.}~\cite{Chen2017} developed a deep neural network within a Bayesian framework to jointly estimate the road surface and its boundaries. In \cite{CaltagironeEtAl2017}, LIDAR point clouds are transformed into 2D top-view images that are then used as input for an FCN to carry out road segmentation. Shinzato \textit{et al.} \cite{ShinzatoEtAl2014} instead projected the point clouds onto the camera image plane and then used a graph-based approach to discriminate between obstacle and obstacle-free space.
	Some methods, such as those found in \cite{XiaoEtAl2015} and \cite{XiaoEtAl2017}, tackled road detection by performing LIDAR-camera fusion within the framework of conditional random fields (CRFs).
	
	Eitel \textit{et al.}~\cite{EitelEtAl2015} proposed to carry out objection recognition by fusing depth maps and color images with a CNN. In \cite{Schlosser2016}, LIDAR point clouds were transformed into their HHA (horizontal disparity, height above the ground, and angle) representation \cite{GuptaEtAl2014} and then combined with RGB images using a variety of CNN fusion strategies for performing pedestrian detection. More recently, Asvadi \textit{et al.}~\cite{AsvadiEtAl2017} developed a system for vehicle detection  that integrates LIDAR and color camera data within a deep learning framework.
	
	Investigating another line of research, in \cite{Bellone2018} a support vector machine (SVM) to carry out road detection on 3D cloud data in challenging scenarios. Using SVM, Zhou \textit{et al.}~\cite{Zhou10}, built a road detection algorithm enabling on-line learning, meaning that this method is able to update the training data, thus reducing the probability of misclassification.  
	Moreover, in more recent research the task of road detection has been extended to challenging scenarios such as slippery roads and adverse weather \cite{zhao2017road}.

\section{Network architectures}
\label{sec:deepstuff}
\subsection{Base FCN} 
\label{sec:base_fcn}
The base neural network used in this work consists of a fully convolutional encoder-decoder that also contains an intermediate context module. This type of architecture has been successfully used in previous publications, such as \cite{CaltagironeEtAl2017} and \cite{CaltagironeEtAlITSC2017}, and it is illustrated in Fig.~\ref{fig:network}. The encoder consists of 5 convolutional layers: $4\times4$ convolutions with stride 2 are used in order to downsample the input tensors thus reducing memory requirements. The context module consists of 9 convolutional layers with $3\times3$ kernels and exponentially growing dilation \cite{YuEtAl2015}. This makes it possible to quickly grow the network's receptive field  while limiting the number of layers. A large receptive field is beneficial for aggregating information within a large region of the input. More details about the context module are provided in Table \ref{tab:nets_properties}. The decoder contains 6 convolutional layers and its purpose is to further process and upsample the input tensors. Upsampling is achieved by using 3 strided convolutional layers with $4\times4$ kernels and stride 2. Each convolutional layer is followed~by an \textit{exponential linear unit} (ELU) layer \cite{ClevertEtAl2015} which implements the following function:
	\[ f(x) =
	\begin{cases}
	x      & \quad \text{if } x \geq 0\\
	\mathrm{e}^{x} - 1  & \quad \text{otherwise} 
	\end{cases}
	\]
	For regularization, spatial dropout layers, with dropout probability $p=0.25$, have been added after each convolutional layer within the context module. This means that, during training, each feature map of a given convolutional layer has a probability $p$ of being set to zero. This technique was shown to be more effective \cite{tompson2015efficient} than the original dropout implementation \cite{Srivastava2014dropout} for improving generalization performance.

\begin{table}[h]
	\centering
	\caption{Context module architecture. The context module consists of 9 convolutional layers with exponentially growing dilation factor. Each convolutional layer is followed by a spatial dropout layer with $p=0.25$. Zero-padding is applied throughout the context module in order to preserve the width and height of the feature maps.} 
	\label{tab:nets_properties}
	\resizebox{\columnwidth}{!}{%	
		\begin{tabular}{|c|c|c|c|c|c|c|c|c|c|c|c|}
			\hline
			\multicolumn{1}{|c|}{Layer}&\multicolumn{1}{c|}{6}&\multicolumn{1}{c|}{7}&\multicolumn{1}{c|}{8}&\multicolumn{1}{c|}{9}&\multicolumn{1}{c|}{10}&\multicolumn{1}{c|}{11}&\multicolumn{1}{c|}{12} &\multicolumn{1}{c|}{13} &\multicolumn{1}{c|}{14}\\
			\hline
			Dilation H & 1 & 1 & 1 & 2 & 4 & 8 & 16 & 1 & - \\
			Receptive field H & 3 & 5 & 7 & 11 & 19 & 35 & 67 & 69 & 69\\
			\hline
			Dilation W & 1 & 1 & 2 & 4 & 8 & 16 & 32 & 1 & -\\
			Receptive field W & 3  & 5 & 9 & 17 & 33 & 65 & 129 & 131 & 131\\
			\hline
			\# Feature maps & 128 & 128 & 128 & 128 & 128 & 128 & 128 & 128 & 128\\
			Filter size & 3 & 3 & 3 & 3 & 3 & 3 & 3 & 3 & 1\\
			\hline
		\end{tabular}}
	\end{table}	
		
\begin{figure*}[h]
		\centering
		\includegraphics[width=\textwidth]{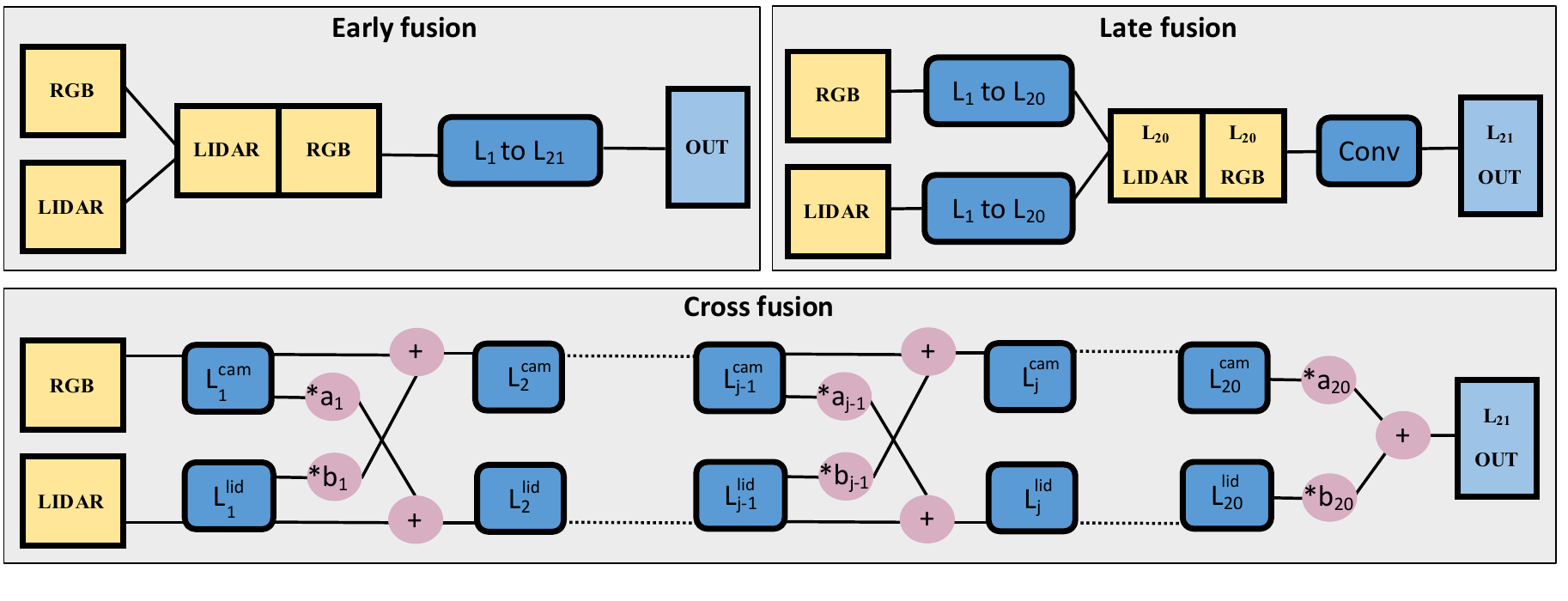}
		\caption{Fusion strategies considered in this work. 1) Early fusion. In this case, the input camera and LIDAR images are concatenated in the depth dimension thus producing a tensor of size $6\times H \times W$. This input tensor is then processed using the base FCN described in Sect.~\ref{sec:base_fcn}. 2) Late fusion. Two parallel streams process the LIDAR and RGB images independently until layer 20. The outputs of $L_{20}$ are then concatenated in the depth dimension and finally fed through a convolutional layer that carries out high-level information fusion. 3) Cross fusion. Also in this case there are two processing branches that, however, are connected by trainable scalar cross connections, $a_{j}$ and $b_{j}$ with $j \in \{1, \ldots, 20\}$. The inputs of each layer, at a given depth, are calculated according to the illustrated computational operations.} 
		\label{fig:fusion_strategies}
	\end{figure*}

\subsection{Early and late fusion}
\label{sec:fusion_strategies}
	This work addresses the task of integrating information acquired with two sensors, an RGB camera and a rotating LIDAR.
	As will be explained in detail in Sect.~\ref{sec:datapreprocessing}, the LIDAR point clouds are transformed into a set of 2D images (in the following denoted as ZYX) that have the same spatial size as the camera images. Given this setup, the integration of camera and LIDAR data can be carried out in a straightforward manner using well-known CNN fusion strategies such as \textit{early} and \textit{late fusion} (see, for example, \cite{EitelEtAl2015} and \cite{Schlosser2016}).

	In the early fusion approach, the input LIDAR and camera tensors are simply concatenated in the depth dimension thus producing a tensor with 6 channels (RGBZYX). This tensor then becomes the input for the base FCN described in Sect.~\ref{sec:base_fcn} which has to learn, from the very beginning, features that combine both sensing modalities; in this case, fusion happens at a very low abstraction level. A graphical illustration of this strategy is presented in Panel 1 of Fig.~\ref{fig:fusion_strategies}.
	
	At the other side of the spectrum is the late fusion. Here, the integration of LIDAR and camera information is carried out at the very end of two independent processing branches, as illustrated in Panel 2 of Fig.~\ref{fig:fusion_strategies}. In this case, fusion happens at the decision level.
	
	A drawback of those approaches is that the developer has to manually decide at which stage the fusion should be done. Here, instead, a novel fusion strategy (\textit{cross fusion}) has been introduced such that the FCN can learn from the data itself, during the training process, where fusion is necessary and to what extent. 
	
	\subsection{Cross fusion}
	The approach proposed in this work is represented in Panel 3 of Fig.~\ref{fig:fusion_strategies}. The rationale behind this strategy is to allow the FCN to integrate information at any processing depth instead of limiting it to a single level, as was the case in the previously mentioned methods.
	For example, the input tensors at depth $j$, denoted as $I_{j}^{\text{Cam}}$ and $I_{j}^{\text{Lid}}$, that are fed to layers $L_{j}^{\text{Cam}}$ and $L_{j}^{\text{Lid}}$, respectively, are given by the following expressions:
	\begin{equation}
		I_{j}^{\text{Lid}} = L_{j-1}^{\text{Lid}} + a_{j-1}L_{j-1}^{\text{Cam}}
	\end{equation}
	\begin{equation}
		I_{j}^{\text{Cam}} = L_{j-1}^{\text{Cam}} + b_{j-1}L_{j-1}^{\text{Lid}}
	\end{equation}

\noindent where $a_{j}, b_{j}\in\mathbf{R}$ with $j \in \{1, \ldots, 20\}$ are trainable \textit{cross fusion parameters}. 
The cross fusion parameters are initialized to zero which corresponds to the case of no information flow between the two processing branches. Afterwards, during training, these parameters are adjusted automatically in order to integrate the two information modalities.
		
	\section{Data preprocessing}
	\label{sec:datapreprocessing}
	\noindent In this work, each LIDAR point cloud is converted to a set of three 2D images that make it straightforward to establish correspondences between color intensities and 3D information. Structuring a 3D point cloud in this manner is also convenient for the purpose of using the CNN machinery originally developed for processing camera images.
	
A point cloud acquired with a Velodyne HDL64 consists of approximately 100k points where each point $p$ is specified by its spatial coordinates in the LIDAR coordinate system, that is $p = [x, y, z, 1]^{\text{T}}$. Given the LIDAR-camera transformation matrix $\textbf{T}$, the rectification matrix $\textbf{R}$, and the camera projection matrix $\textbf{P}$, one can calculate the column position, $u$, and the row position, $v$, where the projection of $p$ intersects the camera plane:
	\begin{equation}
		\label{Equation:cameraprojection}
		\lambda \hspace{1mm} [u, v, 1]^{\text{T}} = \textbf{P} \hspace{1mm} \textbf{R}\hspace{1mm}\textbf{T}\hspace{1mm} p
	\end{equation}
where $\lambda$ is a scaling factor that is determined by solving System \eqref{Equation:cameraprojection}. The above transformation is applied to every point in the point cloud, while discarding points such that $\lambda<0$ or when $[u, v]$ falls outside the image.

\begin{figure}[h]
	\centering
	\includegraphics[width=\columnwidth]{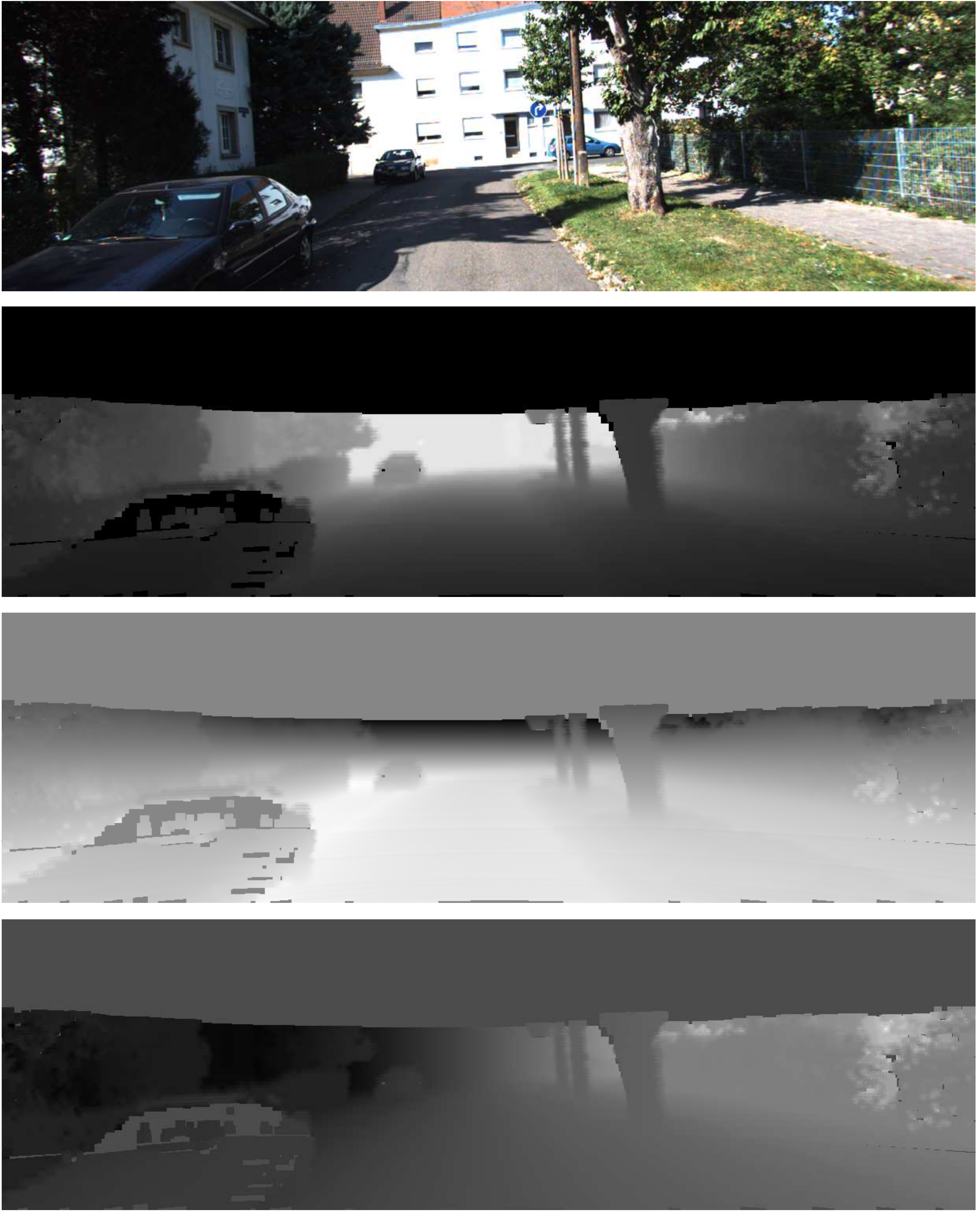}
	\caption{Dense LIDAR images obtained by projecting the point cloud onto the camera plane and then applying the upsampling procedure described in \cite{Premebida2014}. From top to bottom: RGB image, Z channel, Y channel, and X channel. The gray-scale intensities are proportional to the numerical values of the corresponding quantities.}
	\label{Figure:depthupsample}
\end{figure}

Whereas an RGB image contains information about the red, green, and blue intensities of each pixel, the above procedure generates three images, X, Y, and Z where each pixel contains the $x$, $y$, and $z$ coordinates of the 3D point that is projected into it. 
An important difference between camera and LIDAR is that RGB images have valid values for each pixel, whereas, in the LIDAR images, many pixels are set to a default zero value because none of the laser beams hit the corresponding regions. For this reason, it is common practice \cite{Premebida2014,Schlosser2016, Fernandes2014} to upsample the LIDAR images before processing them with machine learning algorithms. This work makes use of the approach introduced by Premebida \textit{et al.} \cite{Premebida2014} to accomplish that. Figure \ref{Figure:depthupsample} shows an example of dense LIDAR images obtained by applying this procedure.

\section{Experiments and Discussion}
\label{sec:experimental_results}
\subsection{Data set}
\label{sec:dataset}
In this work, five different FCNs were considered: ZYX, RGB, Early fusion, Late fusion, and Cross fusion. ZYX denotes the base FCN (see Sect.~\ref{sec:base_fcn}) trained only on LIDAR images. Similarly, RGB is the base FCN trained only on camera images. Early, Late, and Cross fusion are the FCNs implementing the homonymous  fusion strategy (see Sect.~\ref{sec:fusion_strategies}).
Each FCN was trained using exclusively the KITTI road data set which consists of 289 training images and 290 test images taken over several days in city, rural, and highway settings. It is important to mention that most of the training examples were captured in rather ideal weather and lighting conditions, something that might obscure the benefits of combining camera images with additional sensing modalities. For this reason, as will be described in Sect.~\ref{sec:fusion_comparison_validation},
an additional data set of more challenging scenes was included for performance evaluation. Table \ref{tab:kitti_dataset} provides further information regarding the data set splits. Given that the RGB images had different sizes due to the rectification procedure, zero padding was applied to ensure that each training example had the same size of $384\times1248$ pixels. 
\begin{table}[h]
	\centering
	\caption{KITTI road data set: Size and number of images for each category and split. The challenging set was created by the authors and it is not part of the standard KITTI road data set (see Sect.~\ref{sec:fusion_comparison_challenging}).} 
	\label{tab:kitti_dataset}
\resizebox{0.9\columnwidth}{!}{%	
	\begin{tabular}{|c|c|c|c|c|}
		\hline
		\multicolumn{1}{|c|}{Category}&\multicolumn{1}{c|}{Train}&\multicolumn{1}{c|}{Validation}&\multicolumn{1}{c|}{Test}\\
		\hline
		urban marked & 78  & 17 & 96 \\
		urban multiple marked & 80 & 16 & 94 \\
		urban unmarked & 81  & 17 & 100 \\
		\hline
		challenging & - & 33 & - \\
		\hline
	\end{tabular}}
\end{table}

\subsection{Training procedure}
Training was carried out for $N=100$k iterations using the Adam optimization algorithm \cite{KingmaEtAl2014}. The performance of the FCN on the validation set was computed every 1000 iterations and the network's weights were saved if any improvement occurred.
The learning rate $\eta$ was decayed using the poly learning policy \cite{OliveiraEtAl2016} implemented as:
\begin{equation}
	\eta(i) = \eta_{0}\Big(1-\frac{i}{N}\Big)^{\alpha},
\end{equation} 
where $i$ denotes the current iteration number, $\eta_{0}$ is the starting learning rate which was set to $0.0005$, and $\alpha=0.9$. The batch size was set to 1.
Given the small size of the data set, data augmentation was also carried out by applying random rotations in the range $[-20^{\circ}, 20^{\circ}]$ about the center of the images. The FCNs were implemented in PyTorch and trained using an Nvidia GTX1080 GPUs.
The evaluation measures used in the following comparisons are the pixel-wise maximum F-measure (MaxF), precision (PRE), recall (REC), and average precision (AP) \cite{ROC2006}.

\begin{table}[h]
	\caption{Performance comparison of single modality and fusion FCNs evaluated on the validation set.}
	\label{tab:fusion_comparison_validation}
	\centering
\resizebox{\columnwidth}{!}{%	
	\begin{tabular}{|c|c|c|c|c|}
		\hline
		Fusion strategy & \# param.  & MaxF $[\%]$ & PRE $[\%]$ & REC $[\%]$  \\
		%           &  $\%$   & $\%$   & $\%$ & $\%$  & $\%$ & $\%$  \\
		\hline
		ZYX      & 1623395 & 94.96 & 94.05 & 95.89 \\
		RGB      & 1623395 &  96.00 & 96.16 & 95.84 \\
		Early fusion  & 1624931 &  95.41 & 94.62 & 96.21 \\
		Late fusion  & 3246787 &  96.06 & 95.97 & 96.15 \\
		Cross fusion  & 3246830 &  \textbf{96.25} & \textbf{96.17} & \textbf{96.34} \\
		\hline
	\end{tabular}}
\end{table}

\subsection{Comparison of fusion strategies}
\label{sec:fusion_comparison_validation}
The first experiment involved a performance comparison of single modality and fusion FCNs. Table \ref{tab:fusion_comparison_validation} reports the results obtained on the validation set. As can be seen, the overall best performance was achieved by the cross fusion network with a MaxF score of 96.25\%. This is followed by the late fusion FCN that obtained a MaxF score of 96.06\% and then the single modality RGB-FCN  at 96.00\%. The worst performance was obtained by  the FCN that only had access to the LIDAR images resulting in a MaxF score of 94.82\%. This suggests that in scenarios presenting good lighting conditions, camera images are more informative than LIDAR point clouds for the task of road detection. 

\subsection{Challenging scenarios}
\label{sec:fusion_comparison_challenging}
As was mentioned in Sect.~\ref{sec:dataset}, the KITTI road data set mostly consists of examples captured in rather ideal lighting and weather conditions. In those situations, camera images are already, by themselves, quite informative and provide rich discriminative clues for carrying our accurate road detection. This is in part confirmed by noticing that most state-of-the-art algorithms in the KITTI road benchmark are purely camera-based. For this reason, by limiting the evaluation exclusively to the KITTI road data set, it might be difficult to fully appreciate the benefits provided by combining RGB cameras with other sensors, such as LIDARs.
\begin{figure*}[t]
	\centering
	\includegraphics[width=0.9\textwidth]{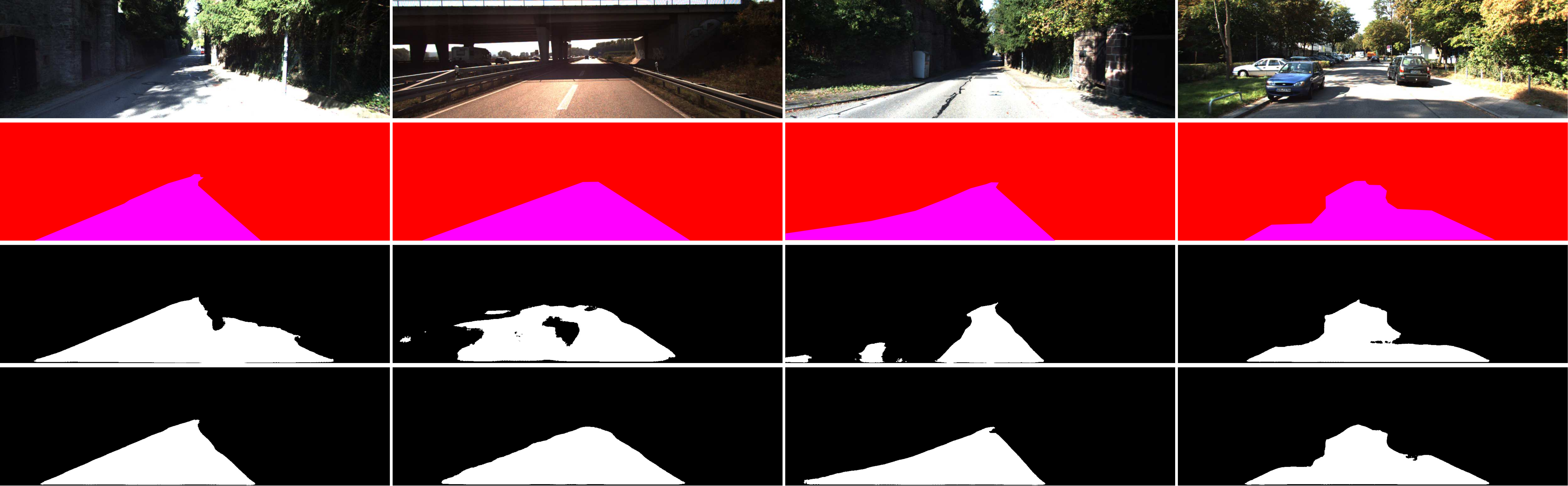}
	\caption{(Top row) Some examples of camera images captured in difficult lighting conditions and included in the challenging set. (Second row) Corresponding ground truth annotations: The road is depicted as violet, whereas not-road is red. (Third row) Road segmentations obtained by the RGB-FCN. (Fourth row) Road segmentations generated by the cross fusion FCN.}
	\label{Figure:comparison_challenging}
\end{figure*}

\begin{figure}[h!]
	\centering
	\includegraphics[width=0.9\columnwidth]{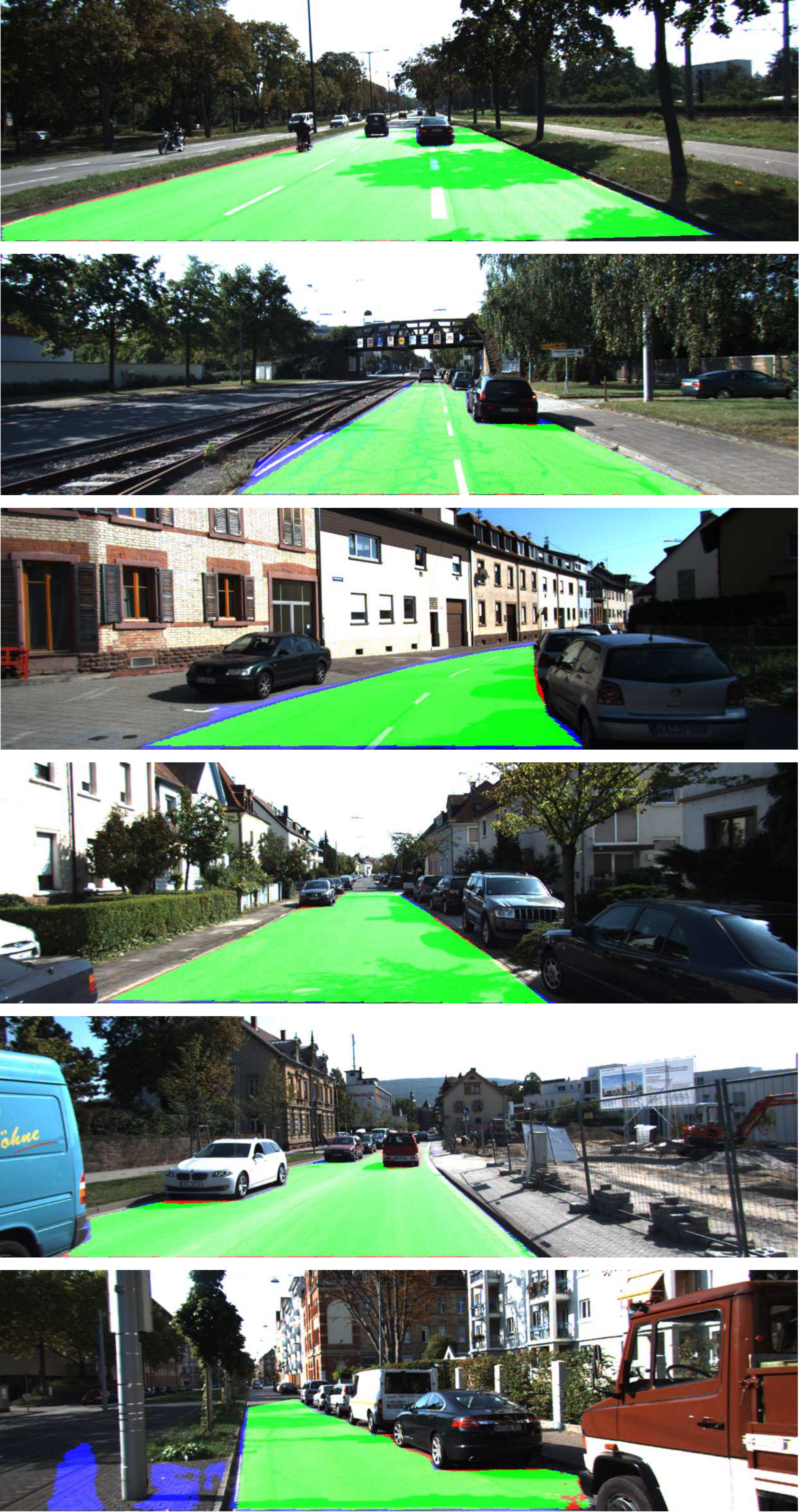}
	\caption{Examples of road segmentations in scenes from the KITTI test set. Correct road classifications are green. Red pixels correspond to false negatives, whereas blue pixels denote false positives.}
	\label{Figure:testres}
\end{figure}

\begin{table}[h]
	\caption{Performance comparison of single modality and fusion FCNs evaluated on the challenging set.}
	\label{tab:fusion_comparison_validation_challenging}
	\centering
\resizebox{\columnwidth}{!}{%	
	\begin{tabular}{|c|c|c|c|c|}
		\hline
		Fusion strategy & \# params  & MaxF $[\%]$ & PRE $[\%]$ & REC $[\%]$  \\
		%           &  $\%$   & $\%$   & $\%$ & $\%$  & $\%$ & $\%$  \\
		\hline
		ZYX      & 1623395 & 95.21 & 93.40 & 97.09 \\
		RGB      & 1623395 &   91.81 & 89.18 & 94.61 \\
		Early fusion  & 1624931 &  95.44 & 93.54 & 97.42 \\
		Late fusion  & 3246787 &   95.24 & 92.73 & 97.09 \\
		Cross fusion  & 3246830 & \textbf{96.02} & \textbf{94.39} & \textbf{97.70} \\
		\hline
	\end{tabular}}
\end{table}

With this consideration in mind, an additional set\footnote{The challenging data set can be found at https://goo.gl/Z5amjQ} consisting of 33 images was extracted from the driving sequences of the KITTI raw data set \cite{GeigerEtAl2013} by looking for scenes that appeared particularly challenging for road segmentation using only the camera sensor, specifically images that presented shadows, strong light reflections, or peculiar lighting conditions affecting the appearance of the road surface. Four such examples and their ground truth annotations are shown in the top two rows of Fig.~\ref{Figure:comparison_challenging}.
The networks trained in Sect.~\ref{sec:fusion_comparison_validation} were also evaluated on this challenging set and their performance is reported in Table \ref{tab:fusion_comparison_validation_challenging}.
As can be noticed, also in this case the cross fusion FCN performed best by achieving a MaxF score of 96.02\%, once more supporting the previous finding that this fusion strategy is more efficient at integrating multimodal information. The RGB-FCN, on the other hand, significantly underperformed, obtaining a MaxF score of 91.81\%. The fusion FCNs and the single modality ZYX-FCN all achieved MaxF scores above 95\%. These results support the intuitive assumption that combining camera images with the spatial information acquired by a LIDAR sensor is indeed beneficial for carrying out road detection in more challenging illumination conditions. Figure \ref{Figure:comparison_challenging} shows some examples of road segmentations obtained with the RGB-FCN (third row) and the cross fusion FCN (fourth row) that qualitatively illustrate the above remark.

\begin{table}[h]
	\centering
	\caption{KITTI road benchmark results (in \%) on the urban road category. Only results of published methods are reported.} 
	\label{table:kitti_results}
\resizebox{\columnwidth}{!}{%	
	\begin{tabular}{|c|c|c|c|c|c|c|}
		\hline
		{Method} & {MaxF} & {AP} & {PRE} & {REC} & {Time (s)}\\		\hline
		\textbf{LidCamNet} (our) & \textbf{96.03} & \textbf{93.93}  & \textbf{96.23}& 95.83 & 0.15\\
		RBNet \cite{Chen2017} &  94.97 & 91.49  & 94.94  & 95.01 & 0.18\\
		StixelNet II \cite{GarnettEtAl2017} &  94.88 & 87.75  & 92.97  & \textbf{96.87} & 1.2\\	
		MultiNet \cite{TeichmannEtAl2016} &  94.88 & 93.71  & 94.84  & 94.91 & 0.17\\	
		LoDNN \cite{CaltagironeEtAl2017} &  94.07 & 92.03  & 92.81  & 95.37 & \textbf{0.018}\\	
		DEEP-DIG \cite{MunozEtAl2017} &  93.98 & 93.65  & 94.26  & 93.69 & 0.14\\			
		Up-Conv-Poly \cite{OliveiraEtAl2016} &  93.83 & 90.47  & 94.00  & 93.67 & 0.08\\
		\hline
	\end{tabular}}
\end{table}

\subsection{KITTI road benchmark}
The cross fusion FCN was also evaluated on the KITTI road benchmark test set. Its performance on the \textit{urban road category} is reported in Table \ref{table:kitti_results} together with the results obtained by other state-of-the-art approaches. At the time of submission, the proposed system was among the best methods in the benchmark. Some examples of road detections on the test set are shown in Fig.~\ref{Figure:testres}. The results on individual categories are reported in Table \ref{table:kitti_results_singlecat}. Additional evaluation metrics and further examples of detections can be found at the KITTI road benchmark website\footnote{https://goo.gl/QNveL1}: The proposed system is called LidCamNet which stands for LIDAR-Camera network. Lastly, several videos of road segmentations on full driving sequences are available at this link https://goo.gl/1oLcmz.

\begin{table}[h]
	\centering
	\caption{KITTI road benchmark results (in \%) on the individual categories. FPR = false positive rate. FNR = false negative rate.} 
	\label{table:kitti_results_singlecat}
\resizebox{\columnwidth}{!}{%	
	\begin{tabular}{|c|c|c|c|c|c|c|}
		\hline
		{Benchmark} & {MaxF} & {AP} & {PRE} & {REC} & {FPR} & {FNR}\\ \hline
		UM\_ROAD & 95.62 & 93.54  & 95.77  & 95.48  & 1.92  & 4.52 \\
		UMM\_ROAD & 97.08  & 95.51  & 97.28  & 96.88  & 2.98  & 3.12 \\
		UU\_ROAD & 94.54  & 92.74  & 94.64  & 94.45  & 1.74  & 5.55 \\
		URBAN\_ROAD & 96.03  & 93.93  & 96.23  & 95.83  & 2.07  & 4.17 \\
		\hline
	\end{tabular}}
\end{table}

\section{Conclusion}
\label{sec:conclusion}
In this paper, a novel fusion FCN has been developed to integrate camera images and LIDAR point clouds for carrying out road detection. Whereas other established fusion strategies found in the literature, such as early and late fusion, are designed to carry out information fusion at a single predefined processing depth, the proposed system incorporates trainable cross connections between the LIDAR and the camera processing branches, in all layers. These connections are initialized to zero, which corresponds to the case of no fusion, and are adjusted during training in order to find a suitable integration level. 

The cross fusion FCN performed best among the single modality and fusion networks considered in this work. Its performance was also evaluated on the KITTI road benchmark where it achieved excellent results, with a MaxF score of $96.03$\% in the urban category, and it is currently among the top-performing algorithms.

An additional data set, consisting of visually challenging examples, was also considered to further highlight the benefits provided by using multiple sensors for carrying out road detection. It was shown that a camera-based FCN that performs quite well in good lighting conditions, will likely underperform in less forgiving situations,  whereas a multimodal system that can leverage the information obtained with a different sensing mechanism can provide more robust and accurate segmentations in a wider spectrum of external conditions.

\section*{Acknowledgment}
The authors gratefully acknowledge financial support from Vinnova/FFI.

\bibliographystyle{IEEEtran} 
\bibliography{references}

\begin{thebibliography}{10}
\providecommand{\url}[1]{#1}
\csname url@samestyle\endcsname
\providecommand{\newblock}{\relax}
\providecommand{\bibinfo}[2]{#2}
\providecommand{\BIBentrySTDinterwordspacing}{\spaceskip=0pt\relax}
\providecommand{\BIBentryALTinterwordstretchfactor}{4}
\providecommand{\BIBentryALTinterwordspacing}{\spaceskip=\fontdimen2\font plus
\BIBentryALTinterwordstretchfactor\fontdimen3\font minus
  \fontdimen4\font\relax}
\providecommand{\BIBforeignlanguage}[2]{{%
\expandafter\ifx\csname l@#1\endcsname\relax
\typeout{** WARNING: IEEEtran.bst: No hyphenation pattern has been}%
\typeout{** loaded for the language `#1'. Using the pattern for}%
\typeout{** the default language instead.}%
\else
\language=\csname l@#1\endcsname
\fi
#2}}
\providecommand{\BIBdecl}{\relax}
\BIBdecl

\bibitem{EvanIEEEspec}
A.~Evan, ``Fatal tesla self-driving car crash reminds us that robots aren't
  perfect,'' in \emph{IEEE Spectrum, Tech. Rep., Jul. 2016. [Online].
  Available:
  https://spectrum.ieee.org/cars-that-think/transportation/self-driving/
  \\fataltesla-autopilot-crash-reminds-us-that-robots-arent-perfect}, 2016.

\bibitem{Endsley17}
M.~R. Endsley, ``Autonomous driving systems: A preliminary naturalistic study
  of the tesla model s,'' \emph{Journal of Cognitive Engineering and Decision
  Making}, vol.~11, no.~3, pp. 225--238, 2017.

\bibitem{Alvarez2007}
J.~M. Alvarez, A.~Lopez, and R.~Baldrich, ``Illuminant-invariant model-based
  road segmentation,'' in \emph{2008 IEEE Intelligent Vehicles Symposium}, June
  2008, pp. 1175--1180.

\bibitem{LecunEtAl2015}
Y.~LeCun, Y.~Bengio, and G.~Hinton, ``Deep learning,'' \emph{Nature}, vol. 521,
  no. 7553, pp. 436--444, 2015.

\bibitem{Broggi95}
A.~Broggi, ``Robust real-time lane and road detection in critical shadow
  conditions,'' in \emph{Proceedings of International Symposium on Computer
  Vision - ISCV}, Nov 1995, pp. 353--358.

\bibitem{Broggi95_2}
M.~Bertozzi and A.~Broggi, ``Gold: a parallel real-time stereo vision system
  for generic obstacle and lane detection,'' \emph{IEEE Transactions on Image
  Processing}, vol.~7, no.~1, pp. 62--81, Jan 1998.

\bibitem{TeichmannEtAl2016}
M.~Teichmann, M.~Weber, M.~Zoellner, R.~Cipolla, and R.~Urtasun, ``Multinet:
  Real-time joint semantic reasoning for autonomous driving,'' \emph{arXiv
  preprint arXiv:1612.07695}, 2016.

\bibitem{Chen2017}
Z.~Chen and Z.~Chen, ``Rbnet: A deep neural network for unified road and road
  boundary detection,'' in \emph{International Conference on Neural Information
  Processing}.\hskip 1em plus 0.5em minus 0.4em\relax Springer, 2017, pp.
  677--687.

\bibitem{CaltagironeEtAl2017}
L.~Caltagirone, S.~Scheidegger, L.~Svensson, and M.~Wahde, ``Fast lidar-based
  road detection using fully convolutional neural networks,'' in
  \emph{Intelligent Vehicles Symposium (IV), 2017 IEEE}.\hskip 1em plus 0.5em
  minus 0.4em\relax IEEE, 2017, pp. 1019--1024.

\bibitem{ShinzatoEtAl2014}
P.~Y. Shinzato, D.~F. Wolf, and C.~Stiller, ``Road terrain detection: Avoiding
  common obstacle detection assumptions using sensor fusion,'' in \emph{2014
  IEEE Intelligent Vehicles Symposium Proceedings}.\hskip 1em plus 0.5em minus
  0.4em\relax IEEE, 2014, pp. 687--692.

\bibitem{XiaoEtAl2015}
L.~Xiao, B.~Dai, D.~Liu, T.~Hu, and T.~Wu, ``Crf based road detection with
  multi-sensor fusion,'' in \emph{2015 IEEE Intelligent Vehicles Symposium
  (IV)}, June 2015, pp. 192--198.

\bibitem{XiaoEtAl2017}
L.~Xiao, R.~Wang, B.~Dai, Y.~Fang, D.~Liu, and T.~Wu, ``Hybrid conditional
  random field based camera-lidar fusion for road detection,''
  \emph{Information Sciences}, 2017.

\bibitem{EitelEtAl2015}
A.~Eitel, J.~T. Springenberg, L.~Spinello, M.~Riedmiller, and W.~Burgard,
  ``Multimodal deep learning for robust rgb-d object recognition,'' in
  \emph{Intelligent Robots and Systems (IROS), 2015 IEEE/RSJ International
  Conference on}.\hskip 1em plus 0.5em minus 0.4em\relax IEEE, 2015, pp.
  681--687.

\bibitem{Schlosser2016}
J.~Schlosser, C.~K. Chow, and Z.~Kira, ``Fusing lidar and images for pedestrian
  detection using convolutional neural networks,'' in \emph{Robotics and
  Automation (ICRA), 2016 IEEE International Conference on}.\hskip 1em plus
  0.5em minus 0.4em\relax IEEE, 2016, pp. 2198--2205.

\bibitem{GuptaEtAl2014}
S.~Gupta, R.~Girshick, P.~Arbel{\'a}ez, and J.~Malik, ``Learning rich features
  from rgb-d images for object detection and segmentation,'' in \emph{European
  Conference on Computer Vision}.\hskip 1em plus 0.5em minus 0.4em\relax
  Springer, 2014, pp. 345--360.

\bibitem{AsvadiEtAl2017}
A.~Asvadi, L.~Garrote, C.~Premebida, P.~Peixoto, and U.~J. Nunes, ``Multimodal
  vehicle detection: fusing 3d-lidar and color camera data,'' \emph{Pattern
  Recognition Letters}, 2017.

\bibitem{Bellone2018}
M.~Bellone, G.~Reina, L.~Caltagirone, and M.~Wahde, ``Learning traversability
  from point clouds in challenging scenarios,'' \emph{IEEE Transactions on
  Intelligent Transportation Systems}, vol.~19, no.~1, pp. 296--305, Jan 2018.

\bibitem{Zhou10}
S.~Zhou, J.~Gong, G.~Xiong, H.~Chen, and K.~Iagnemma, ``Road detection using
  support vector machine based on online learning and evaluation,'' in
  \emph{2010 IEEE Intelligent Vehicles Symposium}, June 2010, pp. 256--261.

\bibitem{zhao2017road}
J.~Zhao, H.~Wu, and L.~Chen, ``Road surface state recognition based on svm
  optimization and image segmentation processing,'' \emph{Journal of Advanced
  Transportation}, vol. 2017, 2017.

\bibitem{CaltagironeEtAlITSC2017}
L.~Caltagirone, M.~Bellone, L.~Svensson, and M.~Wahde, ``Lidar-based driving
  path generation using fully convolutional neural networks,'' in
  \emph{International Conference on Intelligent Transportation Systems (ITSC),
  2017 IEEE}.\hskip 1em plus 0.5em minus 0.4em\relax IEEE, 2017, pp. 573--578.

\bibitem{YuEtAl2015}
F.~Yu and V.~Koltun, ``Multi-scale context aggregation by dilated
  convolutions,'' in \emph{ICLR}, 2016.

\bibitem{ClevertEtAl2015}
D.-A. Clevert, T.~Unterthiner, and S.~Hochreiter, ``Fast and accurate deep
  network learning by exponential linear units (elus),'' \emph{CoRR}, vol.
  abs/1511.07289, 2015.

\bibitem{tompson2015efficient}
J.~Tompson, R.~Goroshin, A.~Jain, Y.~LeCun, and C.~Bregler, ``Efficient object
  localization using convolutional networks,'' in \emph{Proceedings of the IEEE
  Conference on Computer Vision and Pattern Recognition}, 2015, pp. 648--656.

\bibitem{Srivastava2014dropout}
N.~Srivastava, G.~E. Hinton, A.~Krizhevsky, I.~Sutskever, and R.~Salakhutdinov,
  ``Dropout: a simple way to prevent neural networks from overfitting.''
  \emph{Journal of machine learning research}, vol.~15, no.~1, pp. 1929--1958,
  2014.

\bibitem{Premebida2014}
C.~Premebida, J.~Carreira, J.~Batista, and U.~Nunes, ``Pedestrian detection
  combining rgb and dense lidar data,'' in \emph{Intelligent Robots and Systems
  (IROS 2014), 2014 IEEE/RSJ International Conference on}.\hskip 1em plus 0.5em
  minus 0.4em\relax IEEE, 2014, pp. 4112--4117.

\bibitem{Fernandes2014}
R.~Fernandes, C.~Premebida, P.~Peixoto, D.~Wolf, and U.~Nunes, ``Road detection
  using high resolution lidar,'' in \emph{2014 IEEE Vehicle Power and
  Propulsion Conference (VPPC)}, Oct 2014, pp. 1--6.

\bibitem{KingmaEtAl2014}
D.~Kingma and J.~Ba, ``Adam: A method for stochastic optimization,''
  \emph{arXiv preprint arXiv:1412.6980}, 2014.

\bibitem{OliveiraEtAl2016}
G.~L. Oliveira, W.~Burgard, and T.~Brox, ``Efficient deep models for monocular
  road segmentation,'' in \emph{2016 IEEE/RSJ International Conference on
  Intelligent Robots and Systems (IROS)}, Oct 2016, pp. 4885--4891.

\bibitem{ROC2006}
\BIBentryALTinterwordspacing
T.~Fawcett, ``An introduction to roc analysis,'' \emph{Pattern Recognition
  Letters}, vol.~27, no.~8, pp. 861 -- 874, 2006, rOC Analysis in Pattern
  Recognition. [Online]. Available:
  \url{http://www.sciencedirect.com/science/article/pii/S016786550500303X}
\BIBentrySTDinterwordspacing

\bibitem{GeigerEtAl2013}
A.~Geiger, P.~Lenz, C.~Stiller, and R.~Urtasun, ``Vision meets robotics: The
  kitti dataset,'' \emph{The International Journal of Robotics Research},
  vol.~32, no.~11, pp. 1231--1237, 2013.

\bibitem{GarnettEtAl2017}
N.~Garnett, S.~Silberstein, S.~Oron, E.~Fetaya, U.~Verner, A.~Ayash,
  V.~Goldner, R.~Cohen, K.~Horn, and D.~Levi, ``Real-time category-based and
  general obstacle detection for autonomous driving,'' in \emph{Proceedings of
  the IEEE Conference on Computer Vision and Pattern Recognition}, 2017, pp.
  198--205.

\bibitem{MunozEtAl2017}
J.~Munoz-Bulnes, C.~Fernandez, I.~Parra, D.~Fern{\'a}ndez-Llorca, and M.~A.
  Sotelo, ``Deep fully convolutional networks with random data augmentation for
  enhanced generalization in road detection,'' \emph{Workshop on Deep Learning
  for Autonomous Driving on IEEE 20th International Conference on Intelligent
  Transportation Systems}, 2017.

\end{thebibliography}

\end{document}